\documentclass{llncs}
\usepackage[utf8]{inputenc}
\usepackage{xcolor}
\usepackage{color}
\usepackage{graphicx}
\usepackage{amsmath}
\usepackage{amssymb}
\usepackage{url}
\usepackage[binary-units=true]{siunitx}

\newcommand*\samethanks[1][\value{footnote}]{\footnotemark[#1]}

\title{Fully-Automated Analysis of Body Composition from CT in Cancer Patients Using Convolutional Neural Networks}
\author{Christopher P. Bridge\inst{1}\thanks{Equal contribution} \and
Michael Rosenthal\inst{2}\samethanks \and
Bradley Wright\inst{1} \and
Gopal Kotecha\inst{1} \and
Florian Fintelmann\inst{3} \and
Fabian Troschel\inst{3} \and
Nityanand Miskin\inst{2} \and
Khanant Desai\inst{2} \and
William Wrobel\inst{2} \and
Ana Babic\inst{4} \and
Natalia Khalaf\inst{2} \and
Lauren Brais\inst{4} \and
Marisa Welch\inst{4} \and
Caitlin Zellers\inst{4} \and
Neil Tenenholtz\inst{1} \and
Mark Michalski\inst{1} \and
Brian Wolpin\inst{4} \and
Katherine Andriole\inst{1}
}

\institute{MGH and BWH Center for Clinical Data Science, Boston, USA \and Brigham and Women's Hospital, Boston, USA \and Massachusetts General Hospital, Boston, USA \and Dana-Farber Cancer Institute, Boston, USA}

\begin{document}

\maketitle

\begin{abstract}
    The amounts of muscle and fat in a person's body, known as body composition, are correlated with cancer risks, cancer survival, and cardiovascular risk. The current gold standard for measuring body composition requires time-consuming manual segmentation of CT images by an expert reader. In this work, we describe a two-step process to fully automate the analysis of CT body composition using a DenseNet to select the CT slice and U-Net to perform segmentation. We train and test our methods on independent cohorts. Our results show Dice scores (0.95-0.98) and correlation coefficients (R=0.99) that are favorable compared to human readers. These results suggest that fully automated body composition analysis is feasible, which could enable both clinical use and large-scale population studies.

\end{abstract}

\section{Introduction} \label{sec:intro}

Body composition (the amounts of fat and muscle in the body) is associated with important outcomes like cancer risk and survival \cite{Martin2013,Prado2008}. The standard for analysis of body composition is to manually segment the body compartments through a single computed tomography (CT) image at the level of the third lumbar vertebra (L3) \cite{Prado2008}. This approach was shown to strongly correlate with whole-body assessments \cite{Shen2004a,Shen2004b}. Slice selection and manual segmentation by an expert analyst require over 20 minutes per scan in our experience. This time-intensive method has limited the feasibility of population-scale research on body composition.

\begin{figure}[b]
	\includegraphics[width=\textwidth]{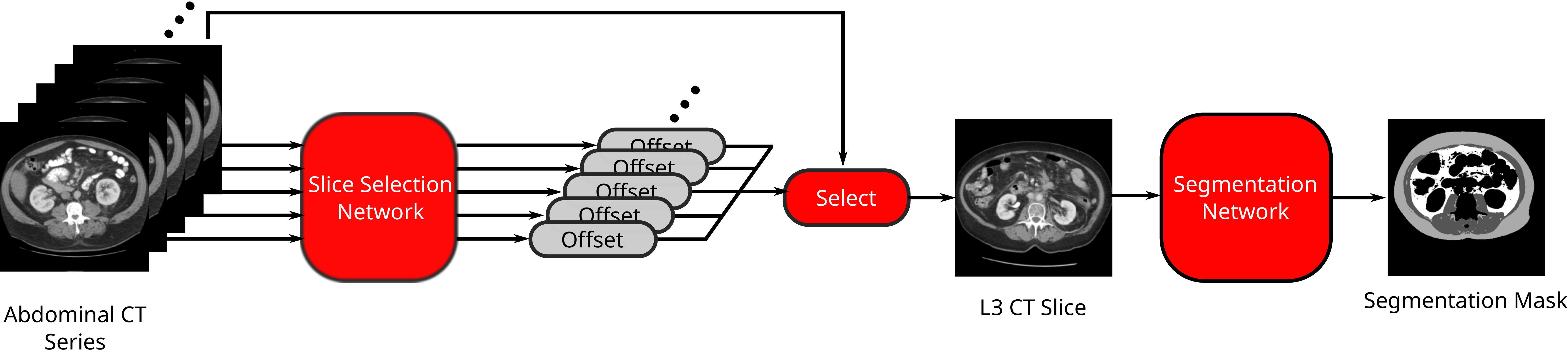}
	\caption{Overview of the body composition workflow}
	\label{fig:workflow-diagram}
\end{figure}

In this paper, we propose a fully automated method to estimate a patient's body composition from an abdominal CT scan and validate the method across two large scale and diverse datasets.
Although automated methods may enable the analysis of muscle and fat distributions across entire abdominal scans, in this work we seek to replicate the gold standard manual approach by segmenting a single CT slice.

Our method therefore breaks the problem of analyzing a CT series down into two steps.
First, a convolutional neural network (CNN) model is used to identify a slice at the L3 level, as described in \S\ref{ssec:slice-selection}.
Next, the chosen slice is passed to a segmentation model (\S\ref{ssec:segmentation}) to estimate the cross-sectional areas of muscle, subcutaneous fat, and visceral fat.
See Figure~\ref{fig:workflow-diagram} for an overview of the workflow.

We demonstrate on a large and diverse dataset that efficient, repeatable and accurate automatic body composition analysis is possible from routinely-acquired CT images (\S\ref{sec:results}).

\section{Related Work}

A number of previous works \cite{Kim2016,Kullberg2017,Lee2017,Parikh2017,Popuri2016} have demonstrated automated methods to segment body fat and/or muscle from axial CT images.
Typically these have depended on handcrafted procedures. In this work, we take a different approach using deep learning methods trained on expert-annotated data.

Of particular note is the work of Popuri et al~\cite{Popuri2016} who use a finite element model (FEM) based approach to segmentation regularised by a statistical deformation model (SDM) prior and achieve high accuracy on a large and diverse dataset, but assume pre-selected slices at the L3 and T4 (thoracic) levels.
They also demonstrate segmentation of muscle and fat, but do not make the clinically significant distinction between visceral fat and subcutaneous fat.

Lee et al.~\cite{Lee2017} previously demonstrated quantification of muscle tissue from single L3 slices using fully convolutional networks.
We improve upon this work by using more modern segmentation architectures, resulting in better performance, and add the ability to segment visceral and subcutaneous fat in order to provide a more comprehensive assessment of body composition.
Furthermore we add a slice-selection step to allow the model to operate on entire CT series without any human intervention, opening up the potential for large-scale cohort analysis.

Belharbi et al.~\cite{Belharbi2017} perform slice selection using convolutional neural networks and use a regression approach on the maximum intensity projection (MIP) image.
We adopt a similar approach and show that the preprocessing to find the MIP is unnecessary and regression based on a single axial slice is highly accurate.

\section{Methodology}
\subsection{Cohorts}

The training cohort (Dataset A) used in this study is composed of 595 CT scans from subjects with biopsy-proven pancreatic adenocarcinoma who were treated at any of several collaborating centers (Brigham and Women's Hospital, Dana-Farber Cancer Institute, and others).
Our group has previously used this cohort to demonstrate that body composition, as determined through manual segmentation through the L3 vertebral body, is associated with overall survival in patients with pancreatic adenocarcinoma \cite{Danai2018}, and that muscle area is associated with outcomes in critical care patients \cite{Foldyna2018}.
Scan parameters, the use of intravenous and oral contrast, and imaging hardware varied widely across the cohort.
All scans were reviewed by a radiologist and a representative slice through L3 was selected.
The three body compartments were manually segmented by trained image analysts using Slice-O-Matic software (Tomovision, Canada).
Standard attenuation constraints were used: -29 to 150 HU for muscle and -190 to -30 HU for fat \cite{Prado2008}.
All segmentations were reviewed and corrected by a board-certified radiologist (MR).
Dataset A was randomly divided into 412 training, 94 validation and 89 segmentation test series.

The testing cohort (Dataset B) is composed of 534 CT scans from subjects with lymphoma treated at a single institution (Massachusetts General Hospital).
Scan parameters and imaging hardware varied across the cohort.
Slice selection and manual segmentation were performed by a trained image analyst and revised by a board-certified radiologist (FF).
Of the total number of series, 512, 473, and 514 series had manual segmentations for muscle, subcutaneous fat, and visceral fat respectively.
Segmentation in this cohort used the same attenuation constraints but was performed in Osirix (Pixmeo, Geneva).
This dataset was used to test the full body composition estimation framework.

\subsection{Slice Selection Model} \label{ssec:slice-selection}

The first step in our method is to automatically identify a slice at the L3 level from the full CT volume to be passed on to the segmentation model.
We pose this problem as a slice-wise regression problem, which operates on each slice of the volume independently, followed by post-processing to choose a single slice.

This allows us to use a more efficient 2D network model and allows us to work on a per DICOM image basis, reducing network complexity and avoiding the need to deal with series with different slice spacings, whilst still considering a slice's local context in the selection process.
The model takes as input a 2D CT slice, downsampled to a $256 \times 256$ image, and learns to predict a single continuous-valued output representing the offset of that slice from the L3 region in the craniocaudal direction.
Instead of directly predicting this offset, we find it advantageous to saturate this value into the range 0 to 1 using a sigmoid function, such that the model learns to focus its discriminatory capability within the area around the L3 region.
If we define the $z$-coordinate of a slice as its location along the craniocaudal axis (the `Slice Location' field in the DICOM metadata) and the $z$-coordinate of the L3 slice in that series is known to be $z_{\text{L3}}$ then the model learns to predict the regression target $r\left( z \right)$ where,

\begin{equation}
	r\left( z \right) = \frac{1}{1 + e^{-\tau \left(z -  z_{\text{L3}}\right)}}
\end{equation}

\noindent and $\tau$ is a free parameter defining the size of the region of interest.
Based on preliminary experiments, we found $\tau = $ \SI{20}{\milli\meter} to be a suitable value.

We experimented with variations on two state-of-the-art CNN architectures: ResNeXt~\cite{Xie2017} and DenseNet~\cite{Huang2017}.
Each of these architectures has recently achieved excellent performance on large-scale image classification tasks and is designed to overcome common problems with training very deep neural networks by introducing skip connections to allow gradients to propagate more directly back to earlier layers of the network.
To adapt the architectures for regression, we replace the final fully-connected layer and softmax activation with a fully-connected layer with a single output unit and a sigmoid activation function in order to output a number in the range 0 to 1.
We then apply a mean absolute error loss between this output and the regression target $r\left( z \right)$.

\begin{figure}[tb]
	\centering
	\includegraphics[width=\textwidth]{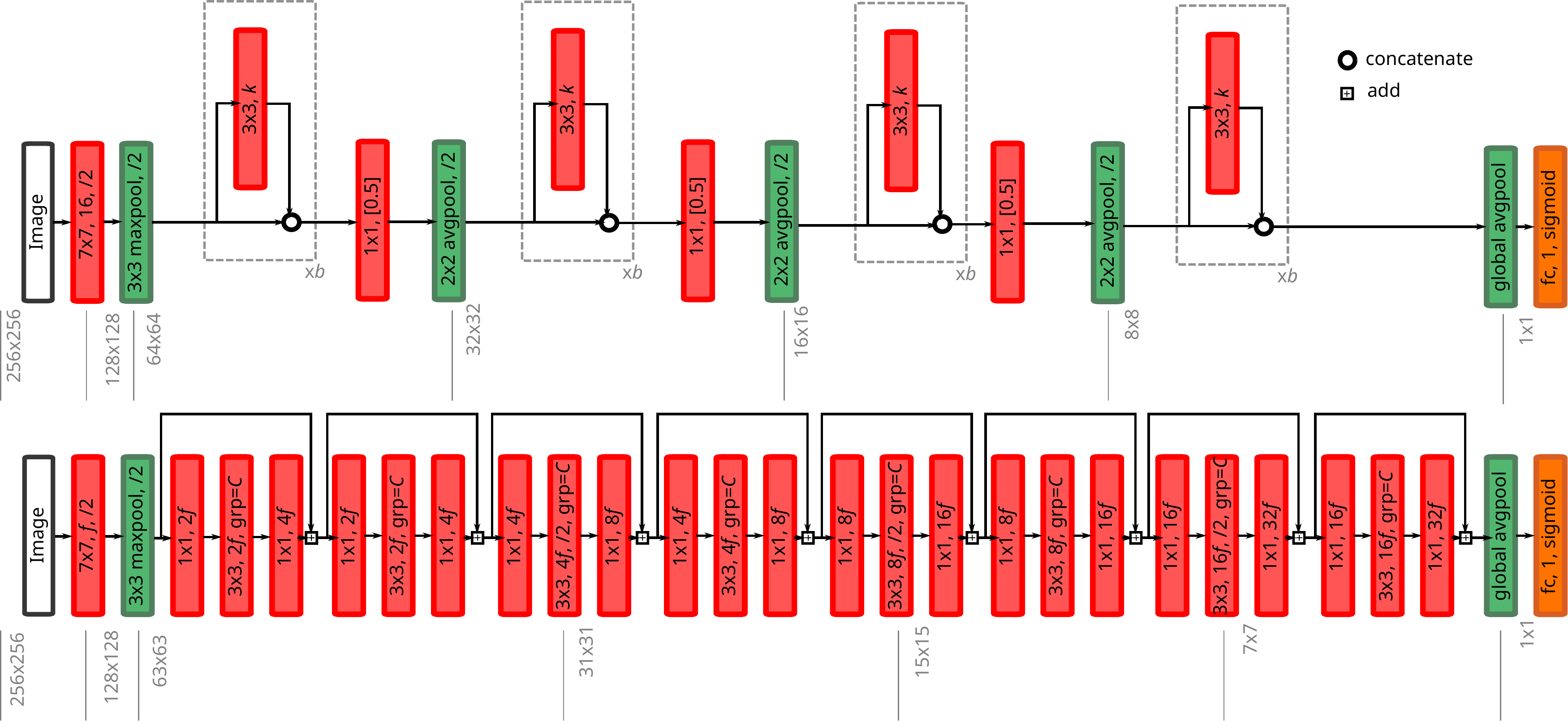}
	\caption{Schematics of networks: \textit{below} ResNeXt, \textit{above} DenseNet. Red blocks indicate convolutional layers described by their kernel dimensions and number of output features. All convolutional layers are followed by a batch normalization layer followed by a rectified linear unit (ReLU) activation. Green blocks represent pooling layers. `$/2$' indicates that the conv/pool layer has  a stride of 2, otherwise the stride is 1. In the ResNeXt model `grp$=C$', indicates that the layer is a grouped convolution layer with a cardinality of $C$. In the DenseNet transition blocks `$[0.5]$' indicates that the number of output features is half the number of input features (a compression factor of 0.5).}
	\label{fig:regression-networks}
\end{figure}

Since these architectures were originally developed for the task of multi-class natural image classification on very large-scale datasets, we experiment with various aspects of the architecture in order to find the optimal design for our purposes.
The model architectures are shown in Figure~\ref{fig:regression-networks}.
For the ResNeXt architecture, we experiment with the initial feature width, $f$, and the cardinality, $C$, of the grouped convolution layers.
For the DenseNet architecture we experiment with the number of layers $b$ in each dense block and the `growth rate' $k$, which is the number of features added by each convolutional layer.

At test time, the full series is passed into the model as a sequence of individual slices..
The predicted offsets are placed into an array using the known slice ordering, and the values are smoothed using a small Gaussian kernel ($\sigma=2$ slices) in order to incoporate local context.
Then the location where this smoothed signal crosses 0.5 (corresponding to $z - z_{\text{L3}} = 0$ during training) is chosen as the L3 slice.
If there are multiple such locations, the slice closest to the head is chosen.

\subsection{Tissue Segmentation Model} \label{ssec:segmentation}

Once a slice has been selected according to the slice selection model (\S\ref{ssec:slice-selection}), the full $512 \times 512$ slice is passed to a segmentation model for body composition analysis.
The segmentation network is based on a U-Net model~\cite{Ronneberger2015}, which has previously proved highly effective in a number of biomedical image segmentation tasks.
We add batch normalization before each activation and change the loss function of the network to be a soft Dice maximization loss~\cite{Milletari2016} in order to deal with class imbalances between three tissue classes and the background class.
The full loss function is defined as the sum of the three soft Dice losses for the three non-background classes (muscle, visceral fat, and subcutaneous fat), i.e.

\begin{equation}
	L = - \sum_{c=1}^{3} \left(  \frac{2\sum_{i=0}^N p_{i,c} \, q_{i,c} + \epsilon}{\sum_{i=0}^N p_{i,c} + \sum_{i=0}^N q_{i,c} + \epsilon} \right)
\end{equation}

\noindent where $p_{i,c} \in [0, 1]$ is the predicted probablility (softmax output) of pixel $i$ belonging to class $c$ and $q_{i,c} \in \lbrace 0, 1 \rbrace$ is the ground truth label for pixel $i$ (1 if pixel $i$ belongs to class $c$, otherwise 0).
$\epsilon$ is a small constant that avoids divide-by-zero problems, and was set to 1 in all experiments.

We experiment with different numbers of downsampling/upsampling modules in the architecture, $d$, the number of convolutional layers per module, $l$, and the initial number of features in the network $f$.

\subsection{Training Details}

All models were trained from scratch using the Keras deep learning library with the Tensorflow backend on Nvidia V100 or P100 GPU hardware.
Input image intensities were windowed at train and test time to match the standard viewing range with the center at 40 HU and a width of 400 HU, and then normalized into the range 0-255.
For all models, the Adam optimizer was used with a batch size of 16 images and the training images (and segmentation masks in the case of the segmentation model) were augmented during training by applying small random translations of up to 0.05 times the image size in both the horizontal and vertical directions, drawn from a uniform distribution, and also small rotations of up to 5 degrees in either direction, also drawn from a uniform distribution.

For the slice selection models, training lasted for 75 epochs with the learning rate initially set to 0.001 and reduced by a factor of 10 at $\frac{1}{2}$ and $\frac{3}{4}$ of the way through the training process.
Every slice from the training set series was used as a single training sample along with its known $r\left( z \right)$ value.

For the segmentation model, training lasted for 100 epochs and the learning rate was initially set to 0.1 with the same decay schedule.
The L3 images from the test set series were used along with their manual segmentation masks.

\section{Experiments and Results} \label{sec:results}

\subsection{Model Selection} \label{ssec:model_selection}

The different model architectures were evaluated through their performance on the validation subset of Dataset A.
These results are shown in Table~\ref{tab:model-selection-results}.
It can be seen that the performance of the ResNeXt and DenseNet models is broadly similar, and that in both cases relatively small models can achieve high accuracy.
It is worth noting that the DenseNet models are typically far smaller (in terms of number of parameters) than the ResNeXt models.
For this reason, we chose the DenseNet model with $l=12$, $k=12$ as the final slice selection model.

It can also be seen that the hyperparameters of the U-Net segmentation model do not significantly effect the results, but there is a weak trend that deeper models with more downsampling and upsampling modules achieve a higher accuracy, probably reflecting the increased capacity of the network to capture global context.
Accordingly, we selected the U-Net model with $d=5$, $l=1$, $f=16$ as the segmentation model for further experiments.

\begin{table}[tb]
	\centering
	\begin{tabular}{|c|c|c|c|c|c|c|c|c|c|c|c|c|c|c|c|}
	\hline
	\textbf{Architecture} & \multicolumn{6}{c|}{\textbf{ResNeXt}} & \multicolumn{9}{c|}{\textbf{DenseNet}} \\ \hline
	$f$ & 16 & 32 & 64 & 16 & 32 & 64 & - & - & - & - & - & - & - & - & - \\ \hline
	$C$ & 1 & 1 & 1 & 32 & 32 & 32 & - & - & - & - & - & - & - & - & - \\ \hline
	$b$ & - & - & - & - & - & - & 6 & 6 & 6 & 12 & 12 & 12 & 18 & 18 & 18 \\ \hline
	$k$ & - & - & - & - & - & - & 12 & 18 & 24 & 12 & 18 & 24 & 12 & 18 & 24 \\ \hline
	Loss ($\times 10^{-2}$) & 2.62 & 2.38 & 2.64 & 2.48 & \textbf{2.25} & 2.53 & 2.67 & 3.04 & 2.68 & 2.28 & 2.64 & 2.75 & 2.66 & 2.62 & 2.31  \\ \hline
\end{tabular}

	\\
	\vspace{1em}
	\begin{tabular}{|c|c|c|c|c|c|c|c|c|c|c|c|c|}
	\hline
	\textbf{Architecture} & \multicolumn{12}{c|}{\textbf{U-Net}} \\ \hline
	$d$ & 4 & 4 & 4 & 4 & 4 & 4 & 5 & 5 & 5 & 5 & 6 & 6 \\ \hline
	$f$ & 16 & 16 & 16 & 32 & 32 & 32 & 16 & 16 & 16 & 32 & 16 & 16 \\ \hline
	$l$ & 1 & 2 & 3 & 1 & 2 & 3 & 1 & 2 & 3 & 1 & 1 & 2 \\ \hline
	Loss  & 2.937 & 2.935 & 2.941 & 2.939 & 2.942 & 2.934 & \textbf{2.944} & 2.938 & 2.938 & 2.942 & 2.937 & 2.942 \\ \hline
\end{tabular}

	\caption{Model selection results for the two tasks on the validation partition of Dataset A. \textit{Above} slice selection using mean absolute error loss, \textit{below} segmentation using the Dice loss (a perfect overlap between all three classes would have a loss of 3).}
	\label{tab:model-selection-results}
\end{table}

\subsection{Test Results}

\begin{figure}[ht]
	\centering
	\begin{tabular}{cb{3.0cm}}
	\includegraphics[width=0.5\textwidth]{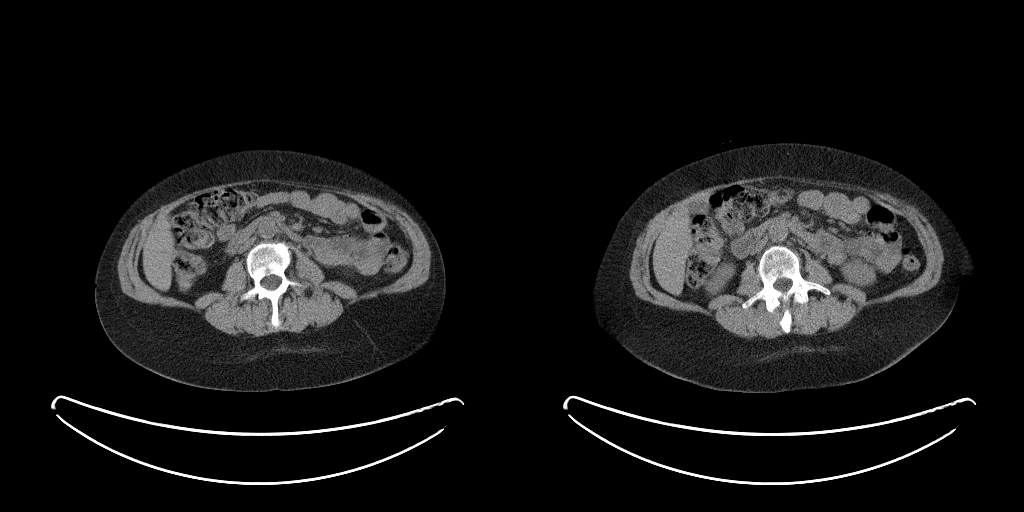}
	\includegraphics[width=0.25\textwidth]{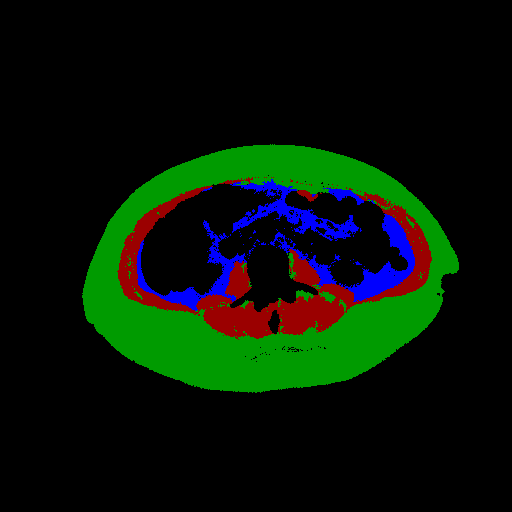}
	& \textbf{Musc.}:~100/\SI{101}{\centi\meter^2} \textbf{Subc.}:~332/\SI{308}{\centi\meter^2} \textbf{Visc.}:~63/\SI{78}{\centi\meter^2}\\
	\includegraphics[width=0.5\textwidth]{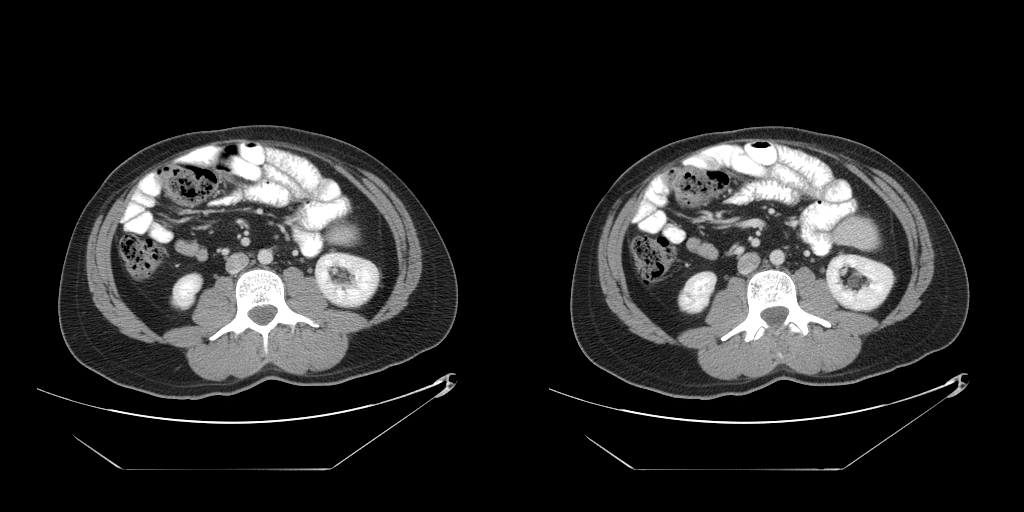}
	\includegraphics[width=0.25\textwidth]{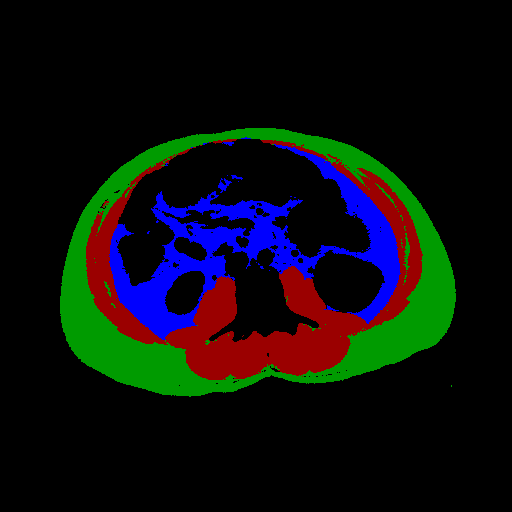}
	& \textbf{Musc.}: 152/\SI{155}{\centi\meter^2} \textbf{Subc.}: 191/\SI{193}{\centi\meter^2} \textbf{Visc.}: 116/\SI{118}{\centi\meter^2}\\
	\includegraphics[width=0.5\textwidth]{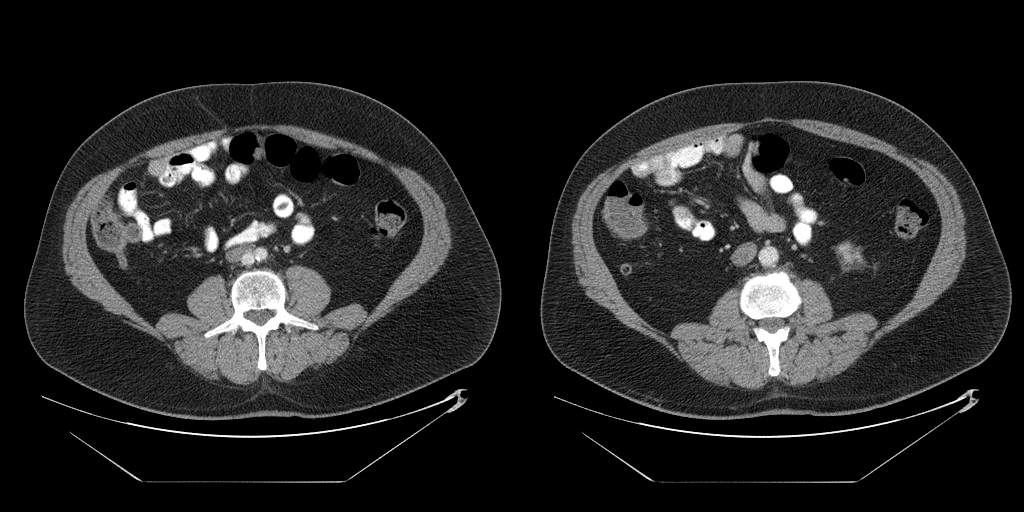}
	\includegraphics[width=0.25\textwidth]{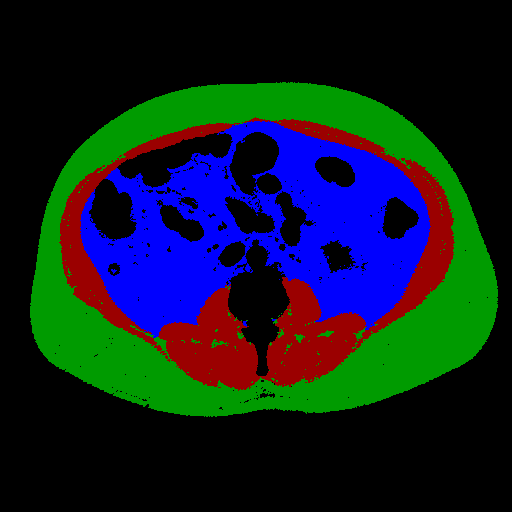}
	& \textbf{Musc.}: 196/\SI{189}{\centi\meter^2} \textbf{Subc.}: 409/\SI{465}{\centi\meter^2} \textbf{Visc.}: 318/\SI{260}{\centi\meter^2}
	\end{tabular}
	\caption{Example results on two randomly-chosen series from the test dataset (Dataset B) and one with poor subcutaneous and visceral fat prediction (third row): \textit{left} manually selected L3 slice, \textit{center} L3 slice chosen by slice selection model, \textit{right} automatically segmented slice showing muscle (red), subcutaneous fat (green) and visceral fat (blue). Areas given in the fourth column are estimated/true values.}
	\label{fig:example_images}
\end{figure}

\begin{figure}[ht]
	\centering
	\includegraphics[width=0.32\textwidth]{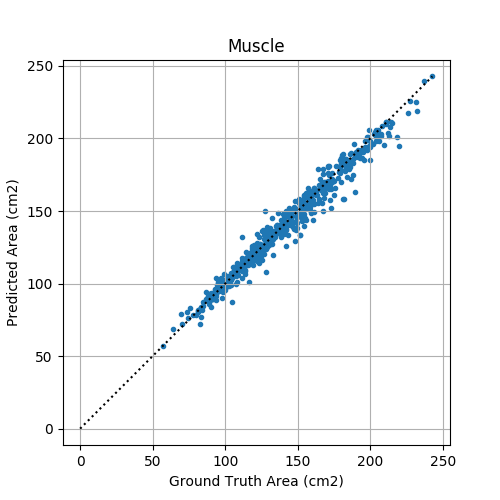}
	\includegraphics[width=0.32\textwidth]{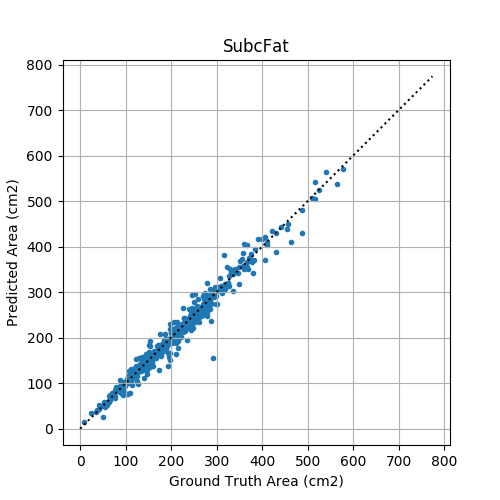}
	\includegraphics[width=0.32\textwidth]{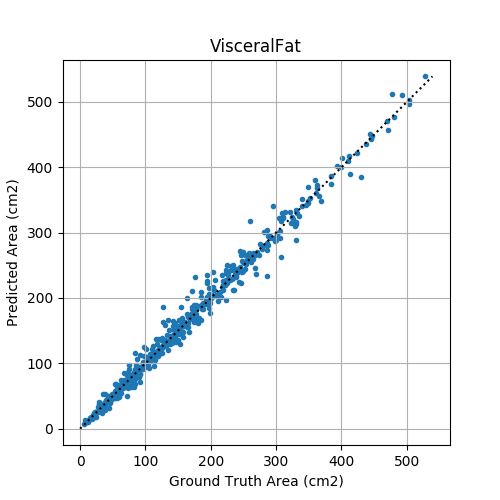}
	\caption{Scatter plots of predicted tissue area versus ground truth tissue area for the three tissue types in Dataset B. Black dashed line indicates perfect prediction ($y=x$).}
	\label{fig:scatterplots}
\end{figure}

Results for the segmentation step in isolation using the selected models on the test partition of Dataset A are shown in Table~\ref{tab:test-results}.
The Dice similarity coefficient (DSC) is used to measure the difference between the automatic segmentation and the manual ground truth.
Our results improve upon those of Lee et al.~\cite{Lee2017}, who had an average DSC of 0.93 for muscle, suggesting that the additional representational power of the U-Net and the more informative three-class training labels were effective at improving network accuracy.
Additionally, our results improve upon those of Popuri et al.~\cite{Popuri2016} who achieved Jaccard indices of 0.904 for muscle, and 0.912 for fat (visceral and subcutaneous as a single class) which correspond to DSC values of 0.950 and 0.954.

The full validation was then performed on Dataset B treating the two models as a single process that takes in a full abdominal CT series and produces estimates of body composition in terms of square cross-sectional area of muscle, subcutaneous fat, and visceral fat.
In this case, the DSC is not an appropriate measure because the segmentation may be performed on a different slice from the ground truth mask.
Table~\ref{tab:test-results} compares the accuracy of the different tissue types and Figure~\ref{fig:example_images} shows some example outputs.

The mean absolute localization error on the Dataset B test set was \SI{9.4}{\milli\meter}, which lies within the range of the L3 vertebra on the majority of patients.


\begin{table}[ht]
	\centering
	\begin{tabular}{|l|c|c|c|}
	\hline
	\textbf{Tissue} & Muscle & Subc.\ Fat & Visc.\ Fat \\ \hline
	Mean DSC & 0.968 & 0.984 & 0.954 \\ \hline
	Std.\ Dev.\ of DSC & 0.034 & 0.021 & 0.100 \\ \hline
\end{tabular}

	\\
	\vspace{1em}
	\begin{tabular}{|c|c|c|c|}
	\hline
	\textbf{Tissue} & Muscle & Subc.\ Fat & Visc.\ Fat \\ \hline
	Mean Absolute Error (\SI{}{\centi\meter^2}) & 4.3 & 10.9 & 7.9 \\ \hline
	Mean Absolute Percentage Error (\%) & 3.1 & 5.9 & 6.5 \\ \hline
	Correlation Coefficient & 0.986 & 0.986 & 0.994 \\ \hline
\end{tabular}

	\caption{Test results on unseen data. \textit{Above} Segmentation results on the test partition of Dataset A, \textit{below} body composition estimation on Dataset B.}
	\label{tab:test-results}
\end{table}

The slice selection model takes approximately \SIrange{0.5}{1.0}{\second} per series to run on our Nvidia V100 GPU hardware, with a further \SIrange{0.02}{0.025}{\second} for the segmentation model.
This compares to the times reported in \cite{Popuri2016} of \SI{0.60}{\second} for the segmentation step alone on a CPU (although our model does make use of GPU hardware).
This makes our approach suitable for use for large scale cohort studies and deployment within a clinical environment.

\section{Conclusions}
We have demonstrated that a two-stage convolutional neural network model can estimate the abdominal muscle and fat areas on abdominal CT scans with high accuracy.
We demonstrate higher segmentation accuracy and faster computation times than the current state of the art.
These findings could enable population-scale research on metabolism by dramatically decreasing the costs associated with this type of analysis.
This could ultimately make routine assessment of body composition a feasible part of the clinical imaging workflow.

\bibliographystyle{splncs03}
\bibliography{references}

\end{document}